\documentclass[conference]{IEEEtran}
\IEEEoverridecommandlockouts
\usepackage{cite}
\usepackage{amsmath,amssymb,amsfonts}
\usepackage{algorithmic}
\usepackage{graphicx}
\usepackage{textcomp}
\usepackage{xcolor}
\usepackage{hyperref}
\usepackage{cleveref}
\usepackage{subcaption}
\def\BibTeX{{\rm B\kern-.05em{\sc i\kern-.025em b}\kern-.08em
    T\kern-.1667em\lower.7ex\hbox{E}\kern-.125emX}}

\def\*#1{\mathbf{#1}}
\usepackage{booktabs}

\newif\iflong
\begin{document}

\title{Exploring the Impact of Disrupted Peer-to-Peer Communications on Fully Decentralized Learning in Disaster Scenarios}

\author{\IEEEauthorblockN{Luigi Palmieri, Chiara Boldrini\IEEEauthorrefmark{1}, Lorenzo Valerio\IEEEauthorrefmark{1}, Andrea Passarella, and Marco Conti}
\IEEEauthorblockA{IIT-CNR, Pisa, Italy\\
% Email: \{name.surname@iit.cnr.it\}}
\{l.palmieri, c.boldrini, l.valerio, a.passarella, m.conti\}@iit.cnr.it}
\thanks{\IEEEauthorrefmark{1} C. Boldrini and L. Valerio contributed equally to this work.}
}

% \author{\IEEEauthorblockN{Luigi Palmieri}
% \IEEEauthorblockA{\textit{IIT-CNR} \\
% % \textit{CNR}\\
% Pisa, Italy \\
% luigi.palmieri@iit.cnr.it}
% \and
% \IEEEauthorblockN{Lorenzo Valerio}
% \IEEEauthorblockA{\textit{IIT-CNR} \\
% % \textit{CNR}\\
% Pisa, Italy \\
% lorenzo.valerio@iit.cnr.it}
% \and
% \IEEEauthorblockN{Chiara Boldrini}
% \IEEEauthorblockA{\textit{IIT-CNR} \\
% % \textit{CNR}\\
% Pisa, Italy \\
% chiara.boldrini@iit.cnr.it}
% \and
% \IEEEauthorblockN{Andrea Passarella}
% \IEEEauthorblockA{\textit{IIT-CNR} \\
% % \textit{CNR}\\
% Pisa, Italy \\
% andrea.passarella@iit.cnr.it}
% \and
% \IEEEauthorblockN{Marco Conti}
% \IEEEauthorblockA{\textit{IIT-CNR} \\
% % \textit{CNR}\\
% Pisa, Italy \\
% marco.conti@iit.cnr.it}
% }

\maketitle

% \IEEEpubidadjcol

\begin{abstract}
Fully decentralized learning enables the distribution of learning resources and decision-making capabilities across multiple user devices or nodes, and is rapidly gaining popularity due to its privacy-preserving and decentralized nature. Importantly, this crowdsourcing of the learning process allows the system to continue functioning even if some nodes are affected or disconnected. In a disaster scenario, communication infrastructure and centralized systems may be disrupted or completely unavailable, hindering the possibility of carrying out standard centralized learning tasks in these settings. Thus, fully decentralized learning can help in this case. However, transitioning from centralized to peer-to-peer communications introduces a dependency between the learning process and the topology of the communication graph among nodes. In a disaster scenario, even peer-to-peer communications are susceptible to abrupt changes, such as devices running out of battery or getting disconnected from others due to their position. In this study, we investigate the effects of various disruptions to peer-to-peer communications on decentralized learning in a disaster setting. We examine the resilience of a decentralized learning process when a subset of devices drop from the process abruptly. To this end, we analyze the difference between losing devices holding data, i.e., potential knowledge, vs. devices contributing only to the graph connectivity, i.e., with no data. Our findings on a Barabasi-Albert graph topology, where training data is distributed across nodes in an IID fashion, indicate that the accuracy of the learning process is more affected by a loss of connectivity than by a loss of data. Nevertheless, the network remains relatively robust, and the learning process can achieve a good level of accuracy.
\end{abstract}


\section{Introduction}
%Definizione e background
The term \emph{network disruption} refers to the occurrence of failures, or disturbances, in the normal functioning of a network, thus resulting in the temporary or prolonged loss of connectivity, reliability, or availability within a network infrastructure.
Network disruptions can be caused by a variety of factors, including hardware or software failures, natural disasters, human error and cyber-attacks.
In our increasingly digital society, the impact of network disruptions can be far-reaching, given the crucial role networks play in communication, data transfer, and access to resources. For example, an organization heavily reliant on network connectivity may experience financial losses, reduced productivity, and damage to its reputation. Industries such as banking, healthcare, transportation, and emergency services heavily rely on networks for their day-to-day operations, making any disruption particularly critical.
Also, individuals heavily rely on networks for communication, access to information, online services, and entertainment. 

In emergency scenarios, infrastructure-based networks (e.g., cellular networks) are often the first to fail or to become so saturated to effectively hinder their use to anyone who is not part of emergency teams. Enabling device-to-device (D2D) -- sometimes referred to as peer-to-peer -- communications has been proposed as a practical solution to cope with this problem~\cite{martin2013evaluating}. Thanks to such a serendipitous D2D connectivity, not only communications among nodes are again possible, but also collaborative tasks can be carried out by the users in a specific area in support of the emergency teams. Collaborative decentralized learning in a disaster scenario is a novel approach that harnesses the power of distributed networks and collective intelligence to tackle the challenges posed by calamities. By leveraging decentralized learning techniques, affected communities can come together, pooling their resources and knowledge, to build resilient solutions. Decentralized learning platforms enable individuals and organizations to share data, insights, and expertise, facilitating real-time analysis and decision-making. This collaborative approach promotes adaptive and inclusive disaster response strategies, as diverse stakeholders contribute their unique perspectives and expertise.

Collaborative decentralized learning that relies exclusively on D2D communications belongs to the class of fully decentralized learning approaches~\cite{lee_opportunistic_2021,palmieri2023effect}. The latter category is an evolution of traditional Federated Learning~\cite{mcmahan_communication-efficient_2017}, where the central parameter server is removed, and user devices collaborate with each other without the need for a central coordinator. This approach is gaining momentum since it combines the privacy-related advantages of Federated Learning with the potentialities of decentralized and uncoordinated optimization and learning. In decentralized collaborative learning, the connectivity between devices is represented with a graph, where an edge between nodes $i$ and $j$ implies that the two nodes can communicate and collaborate. In fact, nodes can only collaborate with their neighbors in the graph, hence the graph topology plays a crucial role in how knowledge flows in such learning schemes. In a decentralized learning scenario, it is realistic to assume that each node only sees a fraction of the overall training data. The learning progresses in communication rounds, where, at each round, a node collects the models of their neighbors, aggregates them, trains the aggregate model on the local data, and then sends the resulting model to their neighbors. 

In disaster scenarios, the user devices themselves can fail, e.g., due to loss of D2D connectivity towards other user devices or because they ran out of battery and cannot promptly recharge. Collaborative decentralized learning is a novel research topic in the distributed learning area, and whether it is robust or not to node failures is yet to be investigated. Understanding the resilience of collaborative decentralized learning to node failure helps us anticipate and mitigate its impact, ensuring the reliability and resilience of critical communication and information systems. Thus, in this paper, we propose a preliminary investigation of
%
% Overall, regardless of the cause, network outages can have serious consequences, disrupting vital services, productivity, and impeding communication channels. Thus, understanding the causes and consequences of network disruptions helps us anticipate and mitigate their impact, ensuring the reliability and resilience of critical communication and information systems. 
%
% In this paper, we investigate 
the impact of disrupted peer-to-peer communications on the ability of a network to retain its learning functionality. We mimic such network failures by switching off some nodes of the network. Since networks based on D2D communications are heavily dependent on the topological structure of the underlying communication graph, we select a realistic graph topology (Barabasi-Albert) and switch off those nodes that act as bridges in the network. Specifically, we consider two cases: one, in which the bridge nodes do not carry data (hence potential knowledge) but only offer connectivity to the rest of the users, and the other one, in which the bridge nodes also carry data that can be exploited for the training process. In the first case, we evaluate the impact of the loss of connectivity only, while in the second case, we test the impact of the joint loss of connectivity and data. The switch-off of bridge nodes is tested at different temporal scales to gauge the impact when bridge nodes are cut off at the beginning vs later on in the learning processes.

The main take-home messages are the following.
\begin{itemize}
    \item Connectivity counts more than data loss: the loss of connectivity has a greater impact on the performance degradation of the system than the joint loss of data;
    % \item The impact of the time of failure is negligible in data-homogeneous settings and scale-free networks:  the time instant when the failure occurs does not significantly impact the final performance of the learning process, i.e., up to 2\% either the failure happens at time 0 or after the system reached 50\% of the final performance accuracy; 
    \item Bridge nodes do have an important role in the learning process: the interaction between these nodes and the worst-performing ones is beneficial for the latter. 
    \item The time of failure mostly affects low-degree nodes: the longer the nodes affected by the disruption stay connected to the network, the higher the final accuracy for the poorly-connected nodes; however, since high-degree nodes preserve their good connectivity despite the switch-off (and perform well with or without the bridge nodes), on average we do not observe a major improvement in the network.
    
\end{itemize}

\section{Related work}
\label{sec:relwork}
As stated in the introduction, network disruption is a central concern in today's interconnected world, whether the network is a communication network, a transportation network, or a social network. As a result, the study of complex network models undergoing network disruption has attracted a lot of attention in the field of network science.
Understanding the vulnerability and resilience of networks is crucial for designing robust systems and developing effective strategies to mitigate potential disruptions. There are different kinds of topologies that can be interesting to study in the context of disaster management. Each topology can have its own strength and weaknesses regarding resilience to failures. For example, in a centralized network topology, all network communication flows through a central node or hub. This topology is particularly vulnerable to disruption because if the central node fails or experiences a disruption, the entire network can become inaccessible or suffer from degraded performance. 

In this context, complex networks, such as the Barabasi-Albert ones, are widely used as reference models due to their ability to capture the inhomogeneous connectivity distribution observed in many real-world networks, thanks to the scale-free property exhibited by this model.
For example, in \cite{albert_error_2000}  the authors investigated the tolerance of complex networks to random errors and targeted attacks. They looked at how the diameter of the networks changed when a small number of nodes were removed, disrupting the interconnectedness of the system, in order to address the networks' fault tolerance. In a related study \cite{holme_attack_2002}, the authors examined the vulnerability of different network models, including Barabasi-Albert graphs, to intentional attacks. Their investigation involved an analysis of how scale-free networks respond to various attack strategies that specifically target nodes based on their properties, such as degree and centrality measures. By examining the behavior and robustness of these networks under different attack scenarios, they provided valuable insights into the dynamics of scale-free networks and their susceptibility to intentional attacks.

Decentralized Learning (DL) extends the typical settings of Federated Learning~\cite{mcmahan_communication-efficient_2017} by removing the existence of the central parameter server. 
% It is gaining momentum since it combines the privacy-related advantages of FL with the potentialities of decentralised and uncoordinated optimisation and learning. 
In~\cite{Roy2019}, a DL framework is introduced for a medical application where multiple hospitals collaborate to train a Neural Network model using their individual and confidential data. Another work~\cite{savazzi_federated_2020} suggests a federated consensus algorithm that extends the FedAvg method proposed by~\cite{mcmahan_communication-efficient_2017} to decentralized environments, primarily focusing on industrial and IoT applications. The work in~\cite{sun_decentralized_2023} proposes a Federated Decentralized Average based on SGD in their research, where they incorporate a momentum term to counterbalance potential drift caused by multiple updates, along with a quantization scheme to reduce communication requirements. The authors of~\cite{palmieri2023effect} study how the network topology affects the learning process in a fully decentralized learning system.

% For example, in \cite{Roy2019}, the authors define a DL framework for a medical application where a number of hospitals collaborate to train a Neural Network model on local and private data. In \cite{savazzi_federated_2020}, the authors propose a federated consensus algorithm extending FedAvg from~\cite{mcmahan_communication-efficient_2017} in decentralised settings, mainly considering industrial and IoT applications. The authors of \cite{sun_decentralized_2023} propose a Federated Decentralised Average based on SGD where the authors include a momentum term to counterbalance the possible drift introduced by the multiple updates and a quantization scheme to reduce communications.

In the context of FL and its robustness, in \cite{kairouz_advances_2021}, the authors provide insights into the challenges and open problems associated with FL, including network disruptions and communication failures. As they pointed out, the FL process involves several steps, including broadcasting a model to the participating clients, local client computation, and the subsequent reporting of client updates to the central aggregator. However, the authors highlight that various system factors can contribute to failures occurring at any of these steps. These failures can range from explicit communication breakdowns to the presence of straggler clients, which experience significant delays in reporting their outputs compared to other nodes in the same communication round. Thus, to optimize efficiency, these straggler clients may be omitted from a communication round, even in the absence of explicit failures. 

In the context of DL, though, the issues of robustness have yet to be tackled. In this work, we present an initial investigation aiming to fill this gap.

% Overall, the research discussed highlights the importance of analyzing network disruption, complex network models, and decentralized federated learning, providing valuable insights into their dynamics, vulnerabilities, and strategies for improving robustness and efficiency. Therefore, there has been increasing interest in exploring the robustness of FL algorithms and specific topologies to network failures. Here, we combine these two main topics by analyzing the impact of disrupted communication flow in Fully Decentralized Learning in a complex network, i.e the Barabasi-Albert network.

\section{Decentralized learning}
\label{sec:decentralized_learning}

We represent the network connecting the nodes as a graph $\mathcal{G}(\mathcal{V},\mathcal{E})$, where $\mathcal{V}$ denotes the set of nodes and $\mathcal{E}$ is the set of edges. 
The decentralized learning algorithms designed for such systems are typically composed of two main blocks: one for the local training of the model using local data and the other one devoted to the exchange and aggregation of the models’ updates. These operations are executed by each node, atomically,  within a single \emph{communication round}.
Each node $i \in \mathcal{V}$ has a local training dataset~$\mathcal{D}_i$ (containing tuples of features and labels $(\*x, y) \in  \mathcal{X} \times \mathcal{Y}$) and a local model $h$ defined by weights $\mathbf{w}_i$, such that $h(\mathbf{x}; \mathbf{w}_i)$ yields the prediction of label $y$ for input $\*x$. Let us denote with $\mathcal{D} = \bigcup_i \mathcal{D}_i$ the set of all data items available in the network and with $\mathcal{P}$ the label distribution in $\mathcal{D}$. In general, $\mathcal{P}_i$ (i.e., the label distribution of the local dataset on node $i$) may be different from $\mathcal{P}$. However, in this paper, since we are interested in investigating how the network disruption might affect the learning process, we choose to limit the degrees of freedom by assuming that (i) the local data are IID across devices and, (ii) all the nodes' models share the same initialization, i.e., at time 0, $\*w_i = \*w_j, \forall i,j$ s.t. $i\neq j$. 

At time 0, the model $h(\cdot; \*w_i)$ is trained on local data by minimizing a target loss function $\ell$: 
\begin{equation}
    \widetilde{\*w}_i = \mathrm{argmin}_{\*w} \sum_{k = 1}^{|\mathcal{D}_i |} \ell(y_k, h(\*x_k;\*w_i)), 
\end{equation}
with $(\*x_k, y_k) \in \mathcal{D}_i$. 
%
% After a number of training steps, the nodes exchange and combine the local models. 
At each communication round, i.e., at each step $t$, a given node $i$ receives the parameters of the local models $\widetilde{\*w}_j$ from its neighbours $j \in \mathcal{N}(i)$ in the social graph and combines them with its local model through the following aggregation policy:
\begin{equation}
    \*w_i^{(t)} \leftarrow \frac{\sum_{j \in \mathcal{N}(i)} \alpha_{ij} \widetilde{\*w}_j^{(t-1)}}{\sum_{j \in \mathcal{N}(i)} A_{ij}},
\end{equation}
where $\mathcal{N}(i)$ is the neighbourhood of node~$i$ including itself, $A_{ij}$ is the $i,j$ element of the adjacency matrix of the graph, and $\alpha_{ij}$ is equal to $\frac{|\mathcal{D}_j|}{\sum_{j \in \mathcal{N}_i} | \mathcal{D}_j|}$ (and captures the relative weight of the local dataset of node $j$ in the neighbourhood of node $i$).
Afterwards, each node begins a new round of local training using the newly aggregated local model. This goes on for a certain number of communication rounds.
% Once the aggregation of models is performed, the new local model is trained again on the local data (in this paper, we use learning rate $\eta$ and momentum $\mu$). 

This strategy, which we denote as DecAvg, extends FedAvg~\cite{mcmahan_communication-efficient_2017} to a decentralized setting and is a generalization of similar strategies proposed in~\cite{sun_decentralized_2023,savazzi_federated_2020}. Differently from the standard Federated Learning, in fully decentralised learning settings (as the ones considered in this paper), the whole process cannot rely on the coordination and supervision of a central entity. This means that, from a node point of view and apart from degenerate cases, the number of other models the node can access to improve its local knowledge is limited to the size of its neighbourhood. Moreover, the connectivity between nodes is a property of the system, and under no circumstances can it be controlled by a node. Conversely, nodes can appear or disappear according to possible failures. 

\section{Evaluation}

\subsection{Experimental settings}
\label{sec:settings}

\subsubsection{Topology of interest}
We consider an unweighted Barabasi-Albert (BA) graph with 100 nodes (shown in Figure~\ref{fig:net_bef_aft}) and a preferential attachment parameter $m = 2$ (meaning that each newly created node starts off with two links, hence 2 becomes the minimum degree for nodes in the network). The topology of the Barabasi-Albert graph is considered to capture important features of real-life networks by incorporating two fundamental principles: preferential attachment (new nodes in the network tend to connect to existing nodes that already have a high number of connections) and organic growth (which aligns with the dynamic nature of many real-world systems). By reproducing these principles, the BA graph model successfully emulates the power-law degree distribution observed in numerous networks, such as social networks, the World Wide Web, and collaboration networks. For this reason, it makes sense to consider it as a realistic topology for our peer-to-peer connectivity graph.

\subsubsection{Selection of nodes that are cut off}
\begin{figure}
\centering
% \vspace{-1cm}
\includegraphics[width = .45\textwidth]{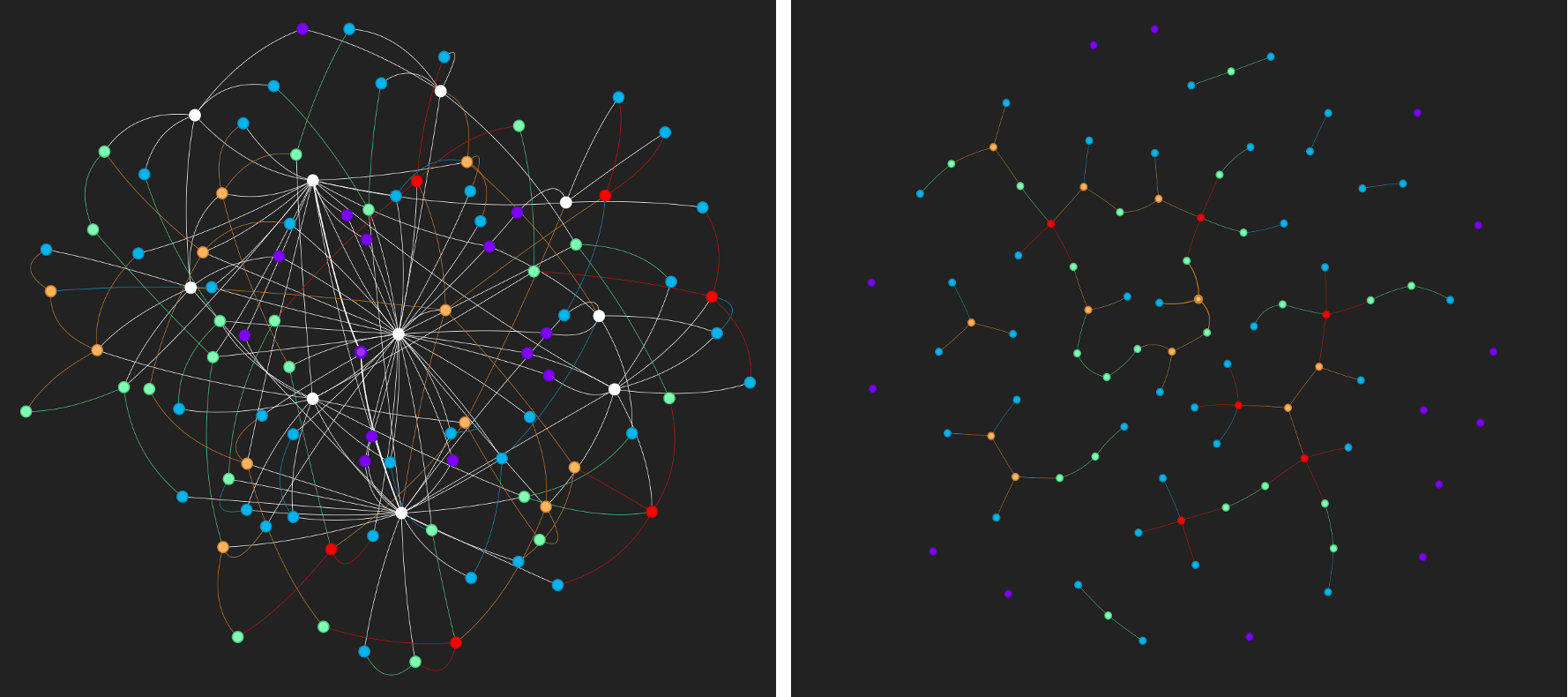}
\captionof{figure}{Our Barabasi-Albert network before (left) and after (right) the disruption. The coloring of nodes corresponds to their degree after the disruption (same legend as \autoref{fig:accuracy_all}). The white nodes in the left image are the nodes that will be switched off at the dropping time.} %\vspace{-10pt}
\label{fig:net_bef_aft}
\end{figure}

Our investigation focuses on examining the effects of disrupted communication in a decentralized learning scenario. Specifically, we introduce disruptions by switching off selected nodes. These nodes are chosen based on their centrality score, which we measure using a metric known as the \emph{structural hole} score~\cite{burt2004structural}. Structural holes refer to the gaps or ``holes" in a social network where there are no direct connections between two or more groups of people. These gaps represent opportunities for actors to act as intermediaries or brokers by connecting otherwise disconnected groups and facilitating the flow of information, resources, or ideas between them. Thus, nodes with higher structural hole scores play a crucial role in facilitating communication across different groups. Here, our strategy consists in switching off the $10\%$ of the total number of nodes that have the highest structural hole score (see \autoref{fig:net_bef_aft}). 

% \begin{figure}
%     \centering
%     \includegraphics[width = .4\textwidth]{ba_100_2_colored.png}\vspace{-10pt}
%     \caption{Barabasi-Albert graph of 100 nodes with preferential attachment $m = 2$. The red nodes are the $10\%$ that is selected based on the structural hole score.} \vspace{-10pt}
%     \label{fig:ba_colored_graph}
% \end{figure}

\subsubsection{Training dataset and data distribution}
For what concerns the training dataset, for our experiments we employ the widely used MNIST image dataset~\cite{lecun1998mnist}. This dataset contains a set of handwritten digits; thus, data are divided into 10 classes (for digits from 0 to 9). The collaborative task undertaken by the nodes is then a standard image classification problem. This problem is relevant to a disaster scenario, where individuals within the affected community can leverage their smartphones or other devices with built-in cameras to capture images of the affected areas, including collapsed buildings, road blockages, or other hazardous conditions.
The 6,000 MNIST training images per class are distributed equally to each node, i.e., we assume that nodes see IID portions of the training data. This allows us to isolate the effects of node disruption from those of data/label distribution. We leave as future work the evaluation of the interplay between the two.

\subsubsection{Disruption analysis}
We focused on two scenarios to investigate network resilience and data resilience. For the first case (Case~1), we investigated network resilience by ensuring that switched-off nodes are not assigned local data. Thus, the highly central nodes will act as bridges but do not contribute knowledge to the learning process. They will just pass along aggregate models without using their local data for training. In the second case (Case~2), the highly central nodes are assigned data, train locally on their own set of data and share their updates with their neighbours. Thus, they are not anymore just a passive element of the network.

Since, in disaster scenarios, disruptions can occur dynamically, here we decided to mimic reality by exploring different values of the communication rounds at which bridge nodes are cut off from the network. For choosing these time instants, we considered the evolution of the accuracy over time (measured in communication rounds) when no node is cut off. Then, we selected $\tau_{drop} = 0, 2, 10$. These different communication rounds represent the initial communication round, the $10\%$, and the $50\%$ of the network's accuracy curve, respectively. Therefore, for each case, we have three different times at which the nodes are switched off. We will refer to them as \emph{dropping times}. 
%Please note that between the two scenarios only the local training changes.

\subsubsection{Other settings}
We implemented the DecAvg scheme within the custom SAISim simulator, available on Zenodo. The simulator is developed in Python and leverages state-of-the-art libraries such as PyTorch and NetworkX for deep learning and complex networks, respectively. It also implements the primitives for supporting fully decentralized learning. The decentralized learning process is run for 200 communication rounds.
For the learning task, we consider a simple classifier as the learned model. The local models of nodes are Multilayer Perceptrons with three layers (sizes 512, 256, 128) and ReLu activation functions. SGD is used for the optimization, with learning
rate 0.01 and momentum 0.5.
We evaluate the impact of the disconnection of bridge nodes by measuring the accuracy obtained by the local models on the standard MNIST test dataset over time. Note that all classes are equally represented in the test set and that the test set is common to all nodes.

\subsection{Results}
In this section, we discuss the impact of bridge nodes being cut off from the network focusing on the two scenarios, Case~1 and Case~2, illustrated in the previous section. Recall that in Case~1, highly central nodes do not have local data, and they only contribute as connecting elements in the graph topology.

\begin{figure*}[p]
    \centering
    \begin{subfigure}{0.8\textwidth}
        \includegraphics[width = \textwidth]{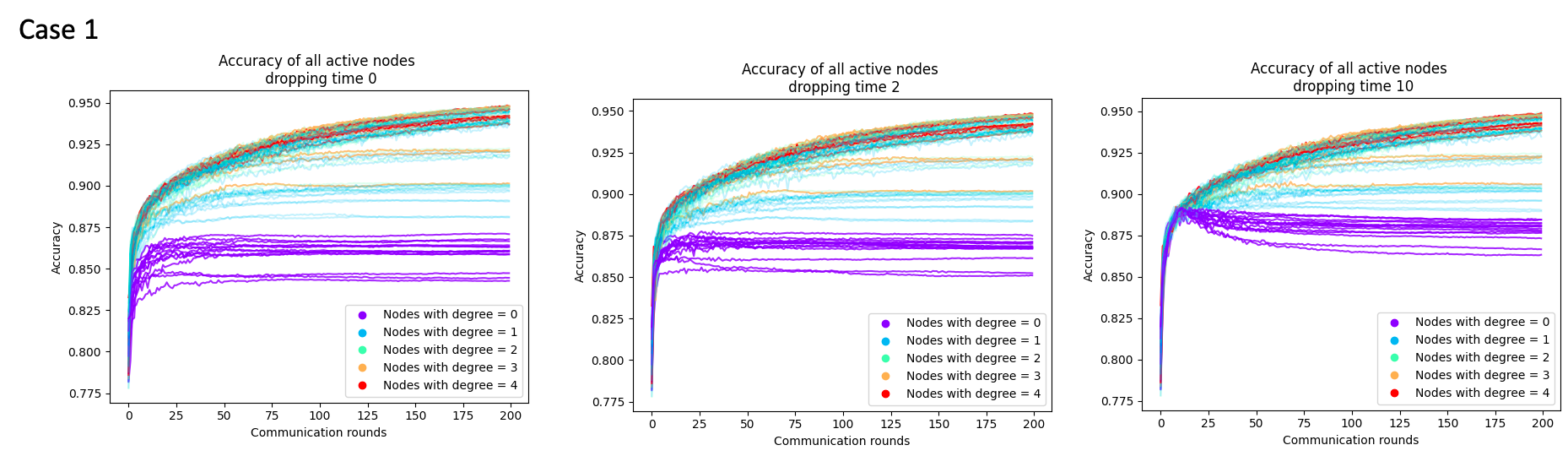}
    \end{subfigure}%
    \vspace{5pt}
    \begin{subfigure}{0.8\textwidth}
        \includegraphics[width = \textwidth]{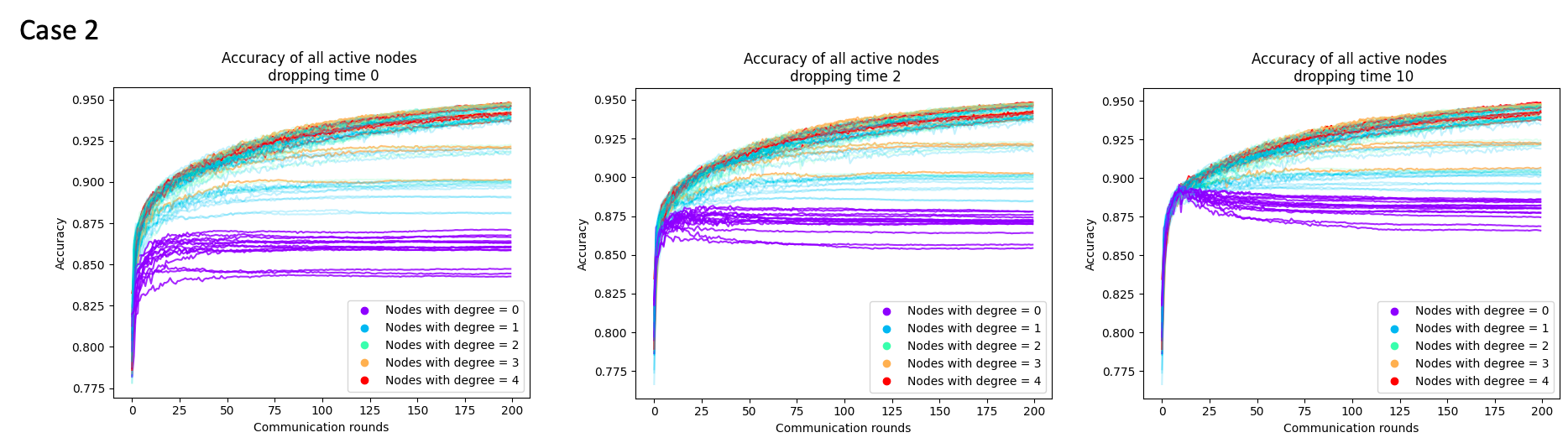}

    \end{subfigure}%
    \caption{Accuracy of all the non-switched off nodes in the two scenarios and for the three different dropping times. From left to right: increasing dropping time. From top to bottom: Case~1 and Case~2. The node degree is computed after the switch-off.} \vspace{-10pt}
    \label{fig:accuracy_all}
\end{figure*}

In~\autoref{fig:accuracy_all}, we show the accuracy of all the active (i.e., non-switched-off) nodes for different dropping times. Note that the degree values shown are calculated based on the disrupted network. 
It can be seen that with increasing dropping time, i.e., going from left to right, the accuracy increases as well, as more and more nodes are able to reach a higher level of accuracy, resulting in a narrower curve beam for both scenarios. This follows naturally after noticing that the central nodes are able to connect the network and, in Case~2, share their information for more communication rounds before getting switched off as the dropping times increase. In all cases, after the dropping time, most curves stop improving and tend to cluster into different groups of similarly performing nodes. Since this happens in both Case~1 and Case~2, it means that this effect is a result of the structural properties of the network. This clusterization can be attributed both to the node degree as well as to the local connectivity of the nodes, which makes it possible for each node to access a certain amount of information from the disrupted network.

Interestingly, in~\autoref{fig:accuracy_all}, with higher values of dropping time, nodes with lower degrees (the worst performing ones) show a degraded accuracy performance that lasts for the subsequent communication rounds. 
% Also, with increasing dropping time, some nodes worsen their performance right after it, resulting in a descendant phase for communication rounds right beyond the dropping time. 
This can be explained by the disrupted line of connection between these nodes and the rest of the network due to the switch-off of the high-centrality nodes. They are not anymore able to get the same amount of information and thus rely much more on their local data, resulting in an overall decrease in their accuracy levels. This is further confirmed by looking at \autoref{fig:net_bef_aft}, where on the right is shown the network after the disruption. The high-degree nodes create chains by linking together a group of nodes. However, the worst-performing nodes (the violet ones) become isolated after the disruption. 

\begin{figure*}[p]
% \vspace{-1cm}
\centering
 \begin{subfigure}{.5\textwidth}
     \includegraphics[width = \textwidth]{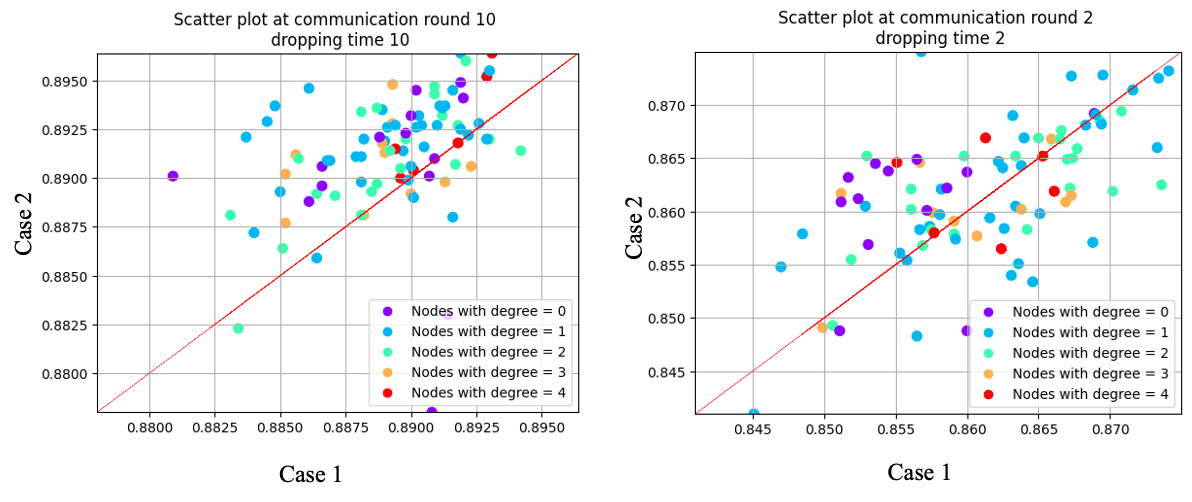}
     \caption{At dropping time}
    \label{subfig:sp_dr}
 \end{subfigure}%
 \hfill
 \begin{subfigure}{.5\textwidth}
     \includegraphics[width =\textwidth]{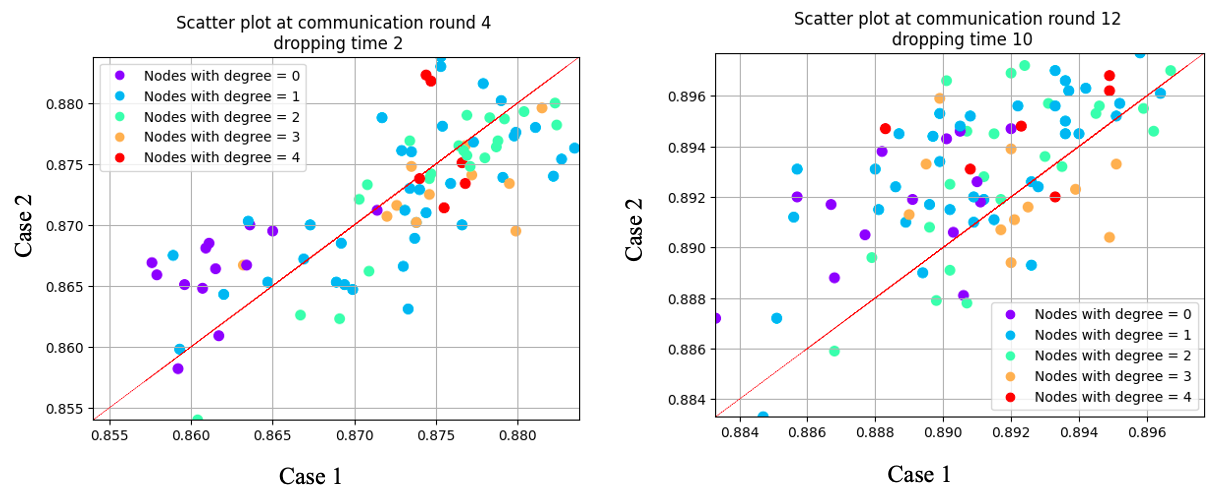}
     \caption{After dropping time}
    \label{subfig:sp_2plus}
 \end{subfigure}%
 \vspace{5pt}
\begin{subfigure}{.5\textwidth}
    \includegraphics[width=\textwidth]{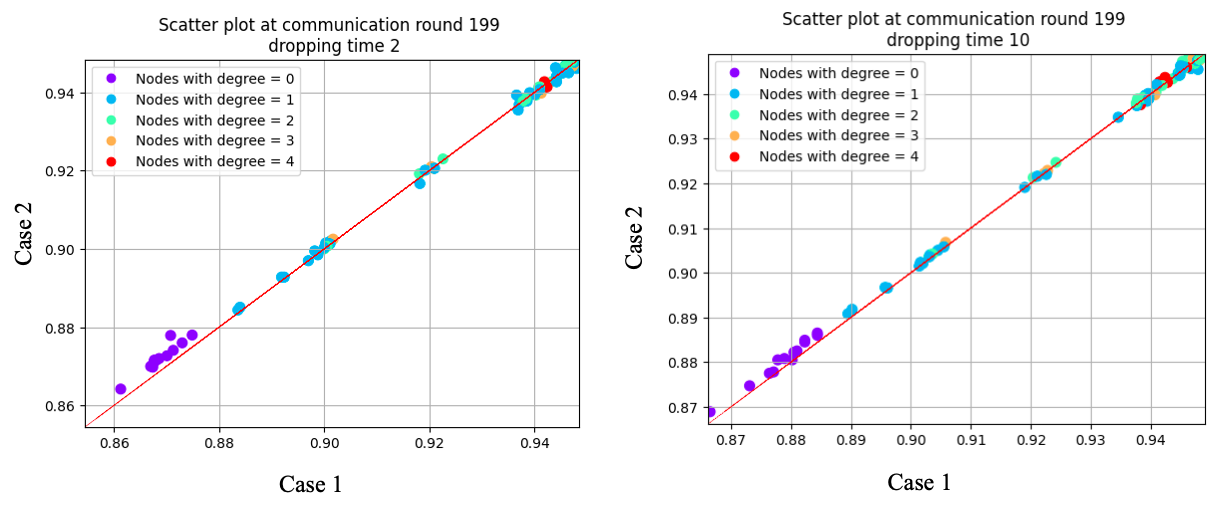}
    \caption{At the last communication round}
    \label{subfig:sp_end}
\end{subfigure}
\caption{Scatterplots of the accuracy in the two cases. The x-axis is the Case~1 accuracy of all the nodes, on the y-axis is the Case~2 one. Each node corresponds to a different point and has a different color as shown in the color gradient bar on the right of each figure. From top to bottom: different values of the communication round at which the scatterplot has been calculated. Identifying by $\tau_{drop}$ the dropping time and $t^*$ the chosen times, then $t^* = [\tau_{drop},\tau_{drop}+2,t_{last}]$, where $t_{last}$ is the last communication round. The red line is the identity line.}
\label{fig:sp_all}
\end{figure*}

%\subsubsection{Case 1 and 2 comparisons}
So far, we have qualitatively compared Case~1 and Case~2. In the following, we compare their accuracies by putting them side by side at significant times during the learning process. This allows us to quantitatively assess the network and data resilience of the considered scenarios. 
%Regarding a comparison between the two cases, i
In \autoref{fig:sp_all}, we show the scatterplots of the accuracy of each node in the two cases for different dropping times and at different communication rounds. Specifically, we take snapshots of the accuracy at exactly the dropping time, two communication rounds after the dropping time ($\tau_{drop}+2)$, and at the last communication round ($t=199$). 
%
% In the first column of \autoref{fig:sp_all}, we observe the effect of a dropping time equal to zero, meaning that the bridge nodes are never contributing to collaborative learning, neither for data nor for connectivity. It is trivial that there is no difference (all the points are on the identity line) between Case~1 and Case~2 when the dropping time $\tau_{drop} = 0$, since they both start without the contribution of bridge nodes.
%
Let us now consider the first row of~\autoref{fig:sp_all}, which contains the accuracy snapshots at the dropping time. We observe that there is no clear winner between Case~1 and Case~2 for $\tau_{drop}=2$, while for $\tau_{drop}=10$ most nodes enjoy a better accuracy in Case~2 than in Case~1. This is due to the fact that bridge nodes in Case~2 contribute data to the learning process as well as connectivity.
Two communication rounds after the dropping time (\autoref{subfig:sp_2plus}), an unexpected situation is observed: several nodes seem to have better accuracy in Case~1 (no data on bridge nodes) than in Case~2 when $\tau_{drop}=2$. The intuitive explanation is that the data contributed by bridge nodes are exposed (via model sharing) to the other nodes for such a short time that they are more a disturbance to the learning process than a help. Vice versa, Case~2 generally keeps its advantage when $\tau_{drop}=10$. Thus, it seems that data needs to stay in the network for some time before being able to contribute to the trained models positively.
Finally, we look at the Case~1 vs Case~2 accuracy at the last communication round (\autoref{subfig:sp_end}). We still see the ``clusters'' of accuracy levels that we observed in~\autoref{fig:accuracy_all}, but now we are able to confirm that these clusters orbit approximately around the same values in both scenarios. 
%
%However, going from left to right, we see that there are differences between Case~1 and Case~2 as we increase the dropping time. Comparing \autoref{subfig:sp_dr} and \autoref{subfig:sp_2plus}, we can see that after the dropping time the accuracy levels start to spread more. That is, until they settle into different clusters as shown in \autoref{subfig:sp_end}. 
At the last communication round, there are some small differences in the accuracy levels reached in the two cases that affect only the worse performing nodes (see \autoref{subfig:sp_end}). Those nodes are above the red line, which means that they perform better in Case~2 than in Case~1. This is because, in contrast to Case~1, in Case~2 the high-centrality nodes have local data and perform local training in between the communication rounds. This, in turn, results in a higher informative update when shared. 
%This is also confirmed by the last column (third figure of \autoref{subfig:sp_dr} and \autoref{subfig:sp_2plus}), where the dots are placed mainly above the red line. Meaning that the nodes consistently perform better when the high-centrality nodes train locally.
%
This also explains why, as we follow nodes with increasing accuracy, the nodes tend to stabilize along the red line (see \autoref{subfig:sp_end}). The nodes that perform better are the ones that can retain a certain amount of connections to the network once the high-centrality nodes are switched off. Therefore, they are less dependent on the bridge nodes, also in the averaging step of the DecAvg update.
In conclusion, from~\autoref{fig:sp_all}, we have learned that worst-performing nodes slightly improve their accuracy when the high-centrality nodes contribute to the process with their local data. However, at convergence, the mean accuracy does not change significantly between the two cases, i.e., the dots adjust on the red line (meaning that the system can compensate for the lost data on the switched-off nodes). Therefore, with IID data distribution and a scale-free network, the decentralised system proves to be pretty robust to the unavailability of some local data, regardless of whether central nodes hold these data.

% \vspace{-3pt}
\section{Conclusions}

In this work, we have focused on collaborative, fully decentralized learning in a disaster scenario, and we have carried out a preliminary investigation on the robustness of this process to the loss of important nodes in the connectivity graph. 
\iflong
In conclusion, our study highlights several important findings regarding the effects of connectivity and data loss in decentralized learning systems. 
\fi
Firstly, we observe that the loss of connectivity after the switch-off affects nodes asymmetrically, with high-degree nodes retaining most of their connections and continuing to learn, while low-degree nodes tend to become disconnected. Additionally, we find that disconnected nodes may continue to experience improvements in accuracy for a few communication rounds after the dropping time, but their performance subsequently plateaus. Interestingly, the accuracy of low-degree nodes may even decrease after the switch-off, indicating that their models struggle to preserve previously acquired knowledge and become overly reliant on local data. The dropping time significantly impacts the final accuracy of poorly connected nodes. 
\iflong
However, as high-degree nodes maintain their strong connectivity regardless of the switch-off and perform well with or without bridge nodes, we observe limited overall improvement in the average network performance. 
\fi
Furthermore, we identify that after the dropping time, the curves of accuracy start to diverge and cluster into distinct groups of similarly performing nodes. Since this happens regardless of whether connectivity-only or connectivity+data are lost, this phenomenon might result from the network's structural properties. Lastly, we note that data loss on bridge nodes predominantly impacts low-degree nodes, as the information provided by bridge nodes is crucial and not easily recoverable for these nodes. These findings shed light on the complex dynamics of decentralized learning systems and emphasize the importance of understanding network structure and connectivity in disaster scenarios.
% \vspace{-5pt}

\section*{Acknowledgment}

This work was partially supported by the H2020 HumaneAI Net (952026) and by the CHIST-ERA-19-XAI010 SAI projects. C. Boldrini and M. Conti's work was partly funded by the Partenariato Esteso PE00000013 - ``FAIR'', A. Passarella's work was partly funded by the Partenariato Esteso PE00000001 - ``RESTART'', both funded by the European Commission under the NextGeneration EU programme, PNRR - M4C2 - Investimento 1.3.

% \vspace{-3pt}

\bibliographystyle{IEEEtran}
\bibliography{references}

% \begin{figure}
% \centering
% % \vspace{-1cm}
% \includegraphics[width = .5\textwidth]{results/network_before_after.png}
% \captionof{figure}{Barabasi-Albert Network before (left) and after (right) the disruption. The coloring of nodes corresponds to the degree-based one used above. The white nodes in the left image are the nodes that will be switched off at the dropping time.} \vspace{-5pt}
% \label{fig:net_bef_aft}
% \end{figure}

\iflong
\begin{figure}
\vspace{-.5cm}
 \includegraphics[width=.5\textwidth]{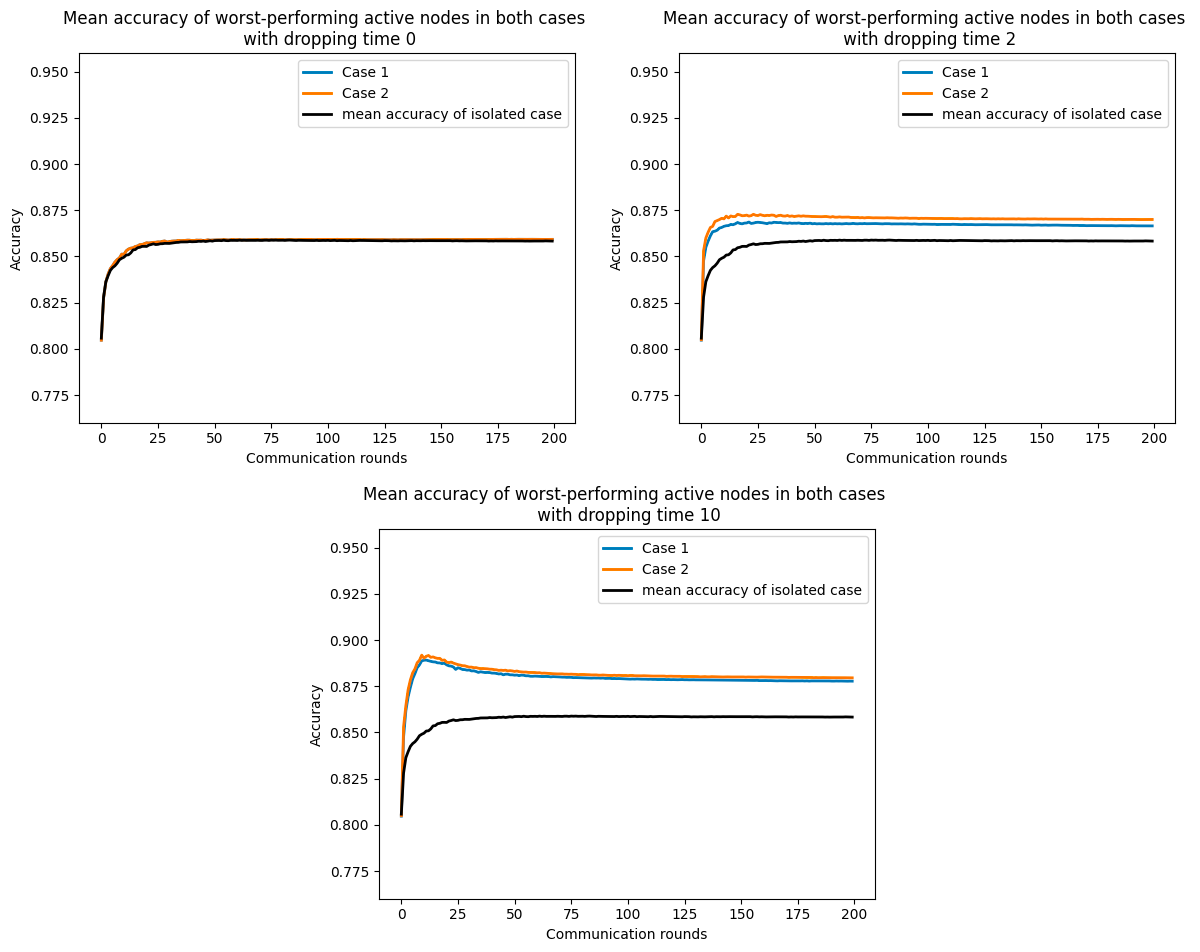}
\captionof{figure}{Mean accuracy comparison of the worst-performing nodes with the one of the isolated case.} \vspace{-10pt}
\label{fig:worst_comparison}
\end{figure}
\vspace{-.7cm}

\begin{figure}[h]
 \centering
    \begin{subfigure} {.24\textwidth}
     \centering
       \includegraphics[width = \textwidth]{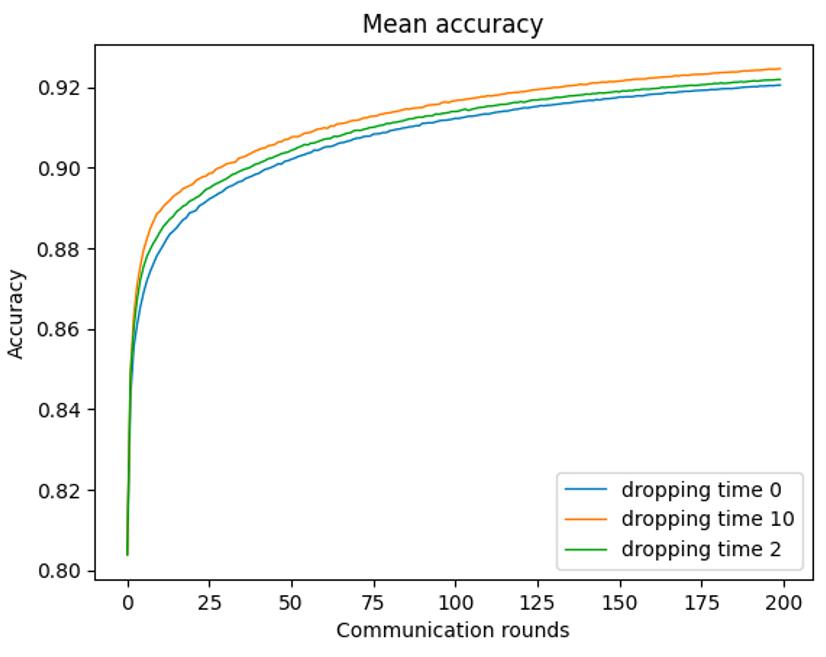}
        \caption{Case1}
        % \label{subfig:sp_dr}
    \end{subfigure}%
    % \hfill
    \begin{subfigure}{.24\textwidth}
     \centering
       \includegraphics[width = \textwidth]{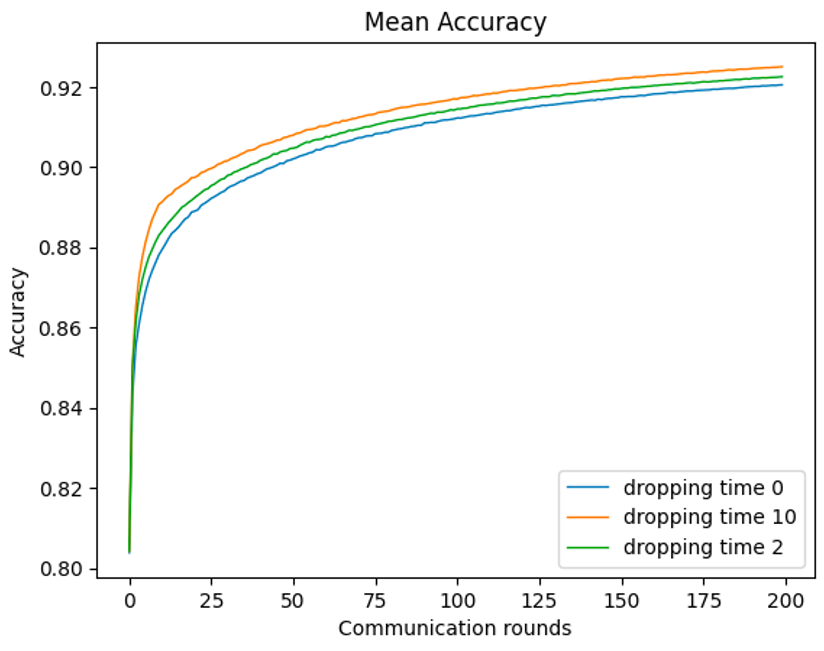}
        \caption{Case2}
        % \label{subfig:sp_2plus}
    \end{subfigure}%
    
    \begin{subfigure}{.24\textwidth}
         \centering
        \includegraphics[width = \textwidth]{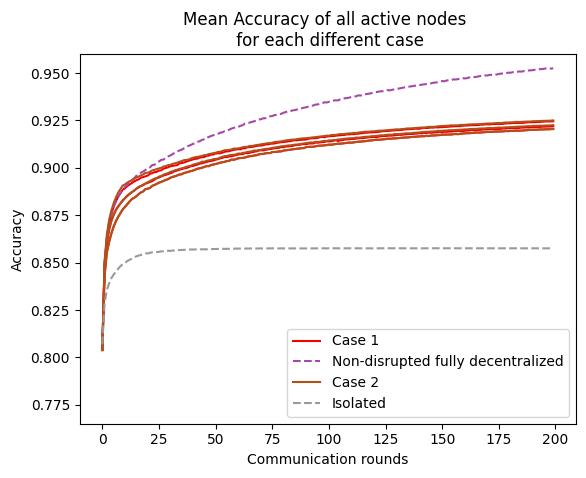}
        \caption{Comparison between all cases.}
        \label{subfig:all_comparison}
    \end{subfigure}
    \caption{Mean accuracy.} \vspace{-20pt}
    \label{fig:case12_mean_acc}
\end{figure}
\fi

\end{document}